\RequirePackage{fix-cm}
\documentclass[twocolumn]{svjour3}          
\smartqed  
\usepackage{graphicx}
\usepackage{multirow}
\usepackage{amsmath}
\usepackage{amssymb}
\usepackage{bbm}
\usepackage{bm}
\usepackage{booktabs}
\usepackage{array} 
\usepackage{blindtext}
\usepackage{comment}
\usepackage{tabularx}

\usepackage{array}

\newcolumntype{P}[1]{>{\centering\arraybackslash}p{#1}}

\usepackage{algorithm}
\usepackage{algorithmic}

\usepackage{subcaption}

\usepackage{xspace}
\usepackage{appendix}
\makeatletter
\DeclareRobustCommand\onedot{\futurelet\@let@token\@onedot}
\def\@onedot{\ifx\@let@token.\else.\null\fi\xspace}

\makeatother

\usepackage{color, colortbl}
\usepackage[table]{xcolor}
\definecolor{remark}{rgb}{1,.5,0} 
\definecolor{citepcolor}{rgb}{0,0.443,0.737} 
\definecolor{linkcolor}{rgb}{0.956,0.298,0.235} 
\definecolor{gray}{gray}{0.95}
\definecolor{cyan}{rgb}{0.831,0.901,0.945}
\definecolor{mygray}{gray}{.9}

\usepackage{url}
\usepackage[misc]{ifsym}

\definecolor{lightgreen}{HTML}{D8ECD1}
\definecolor{edit}{HTML}{C63678}

\usepackage{pifont}

\usepackage{color, colortbl}
\definecolor{citecolor}{HTML}{0071bc}
\definecolor{tabhighlight}{HTML}{e5e5e5}
\usepackage[colorlinks,citecolor=citecolor]{hyperref}

\newcommand{\minisection}[1]{\vspace{0.04in} \noindent {\bf #1}\ \ }

\makeatletter
\renewcommand\paragraph{
  \@startsection{paragraph} 
  {4} 
  {\z@} 
  {.5em \@plus1ex \@minus.2ex} 
  {-.5em} 
  {\normalfont\normalsize\bfseries} 
}
\makeatother
\begin{document}
\sloppy

\title{MaTe3D: Mask-guided Text-based 3D-aware Portrait Editing
}


\author{Kangneng Zhou \and Daiheng Gao \and Xuan Wang\textsuperscript{ \Letter} \and Jie Zhang \and Peng Zhang \and Xusen Sun \and Longhao Zhang \and Shiqi Yang \and Bang Zhang \and Liefeng Bo \and Yaxing Wang\textsuperscript{ \Letter} \and Ming-Ming Cheng
}


\institute{Kangneng Zhou, Yaxing Wang\textsuperscript{ \Letter} and Ming-Ming Cheng \at VCIP, CS, Nankai University \\
              \email{elliszkn@163.com, yaxing@nankai.edu.cn, cmm@nankai.edu.cn}           
           \and
           Daiheng Gao \at
              University of Science and Technology of China
           \and
           Xuan Wang\textsuperscript{ \Letter} \at
              Ant Group \\
              \email{xwang.cv@gmail.com}
           \and
           Jie Zhang \at
              Faculty of Applied Sciences, Macao Polytechnic University
           \and 
            Peng Zhang, Longhao Zhang, Bang Zhang and Liefeng Bo \at
            Institute for Intelligent Computing, Alibaba Group
            \and 
            Xusen Sun \at
            Nanjing University
            \and        
            Shiqi Yang
            \at
            Computer Vision Center, Universitat Aut\`onoma
            \and
            This work was partly done as an intern at Alibaba Group.
}

\date{Received: date / Accepted: date}

\maketitle

\begin{abstract}
3D-aware portrait editing has a wide range of applications in multiple fields. However, current approaches are limited due that they can only perform mask-guided or text-based editing. Even by fusing the two procedures into a model, the editing quality and stability cannot be ensured.
To address this limitation, we propose \textbf{MaTe3D}: mask-guided text-based 3D-aware portrait editing. In this framework, first, we introduce a new SDF-based 3D generator which learns local and global representations with proposed SDF and density consistency losses. This enhances masked-based editing in local areas; 
second, we present a novel distillation strategy: Conditional Distillation on Geometry and Texture (CDGT). Compared to exiting distillation strategies, it mitigates visual ambiguity and avoids mismatch between texture and geometry, thereby producing stable texture and convincing geometry while editing.
Additionally, we create the CatMask-HQ dataset, a large-scale high-resolution cat face annotation for exploration of model generalization and expansion. We perform expensive experiments on both the FFHQ and CatMask-HQ datasets to demonstrate the editing quality and stability of the proposed method. Our method faithfully generates a 3D-aware edited face image based on a modified mask and a text prompt. Our code and models will be publicly at \href{https://github.com/MontaEllis/MaTe3D}{link}.
\keywords{Mask-guided Editing \and Text-guided Editing \and Diffusion Model \and Score Distillation Sampling}
\end{abstract}

\section{Introduction}
\label{intro}
Leveraging the advancements of StyleGAN series~\cite{stylegan},~\cite{stylegan2},~\cite{stylegan3},~\cite{zhou2022sd}, EG3D~\cite{eg3d} and related works like~\cite{stylesdf},~\cite{stylenerf},~\cite{cips3d},~\cite{meshwgan} have successfully created high-quality 3D scenes with consistent views and rich details. Recent 3D-aware image synthesis methods~\cite{fenerf},~\cite{ide3d},~\cite{lenerf} aim to enhance view-consistent image editing capabilities by incorporating additional information like semantic maps. This caters to a range of user needs, including 3D avatar creation, personalized services, and industrial design. 

Existing NeRF-based generation methods with editability can be categorized into two classes: \textit{mask-guided} \cite{fenerf},\cite{ide3d} and \textit{text-guided} \cite{clipnerf},~\cite{lenerf}. Both methods rely on a common pipeline: (1) training a 3D-aware generator; (2) exploring various editing methods based on the well-trained generator. Mask-guided methods aim to facilitate user-friendly editing through hand-drawn sketches or semantic label interfaces. For example, IDE-3D~\cite{ide3d} enables both high-quality and real-time local face control by manipulating a semantic map. However, current mask-guided approaches primarily emphasize shape manipulation and lack texture control capabilities.  On the other hand, text-guided methods strive to produce more expressive results based on textual descriptions by distilling knowledge from large models. LENeRF~\cite{lenerf} utilizes CLIP~\cite{clip} for precise and localized manipulation using text inputs. Nevertheless, existing text-based techniques encounter challenges in accurately controlling shape to achieve desired results(Fig.~\ref{fig:hyperedit}). In this paper, we focus on  using both masked-guided and text-guided 3D image editing with high quality and stability. Masks allow for precise spatial modifications, such as removing or altering specific parts of the scene, while text editing allows for semantic changes based on natural language descriptions. Utilizing both mask and text enhances interactivity and control by enabling users to interact with and modify 3D scenes in more intuitive ways.

To achieve these properties, a simple approach is to combine mask-guided and text-guided techniques presented in the aforementioned part. However, when naively combining the current state-of-the-art methods of both mask-guided and text-based approaches,  the system suffers from \textit{unstable texture} and \textit{unconvincing geometry} (refer to Fig.~\ref{fig:ablation_ckgt}). Also it hinders the  ability to merge visual and textual information, as illustrated in Fig.~\ref{fig:hyperedit}. Unfortunately, there are no prior tailored works exploring how to effectively perform both mask-guided and text-guided manipulation in a single model without conflict.

To address this limitation, we introduce \textit{MaTe3D, Mask-guided Text-based 3D-aware Portrait Editing}, which consists of two steps:(1) modeling 3D generator and (2) effectively using both mask and text information in a single model. Specifically,in the first step we introduce a new SDF-based 3D generator that explicitly models the local representations as well as the global ones. Meanwhile, we also propose the SDF and density based consistency losses to make the generator more accurate. 
The SDF-based 3D generator further enables the possibility of directly constructing semantic probability to learn well-established geometry and view-consistency editing results (Fig.~\ref{fig:dg} and Fig.~\ref{fig:ablation_sdf}).

In the second step, we introduce an innovative editing technique, Condition Distillation on Geometry and Texture (CDGT), tailored for stable texture and convincing geometry while editing. This method leverages a user-provided mask and textual description to facilitate optimal inference, addressing the challenge of 3D mask annotation scarcity through a progressive condition updating scheme. By iteratively refining masks with rendered results from the generator, CDGT improves stable texture, mitigating visual ambiguity across diverse views within the diffusion model. Additionally, our approach employs Score Distillation Sampling (SDS) on the blending of images and their normal maps, ensuring the preservation of the underlying geometry integrity and avoiding mismatch between texture and geometry. Meanwhile, to explore the generalization and expansion of our method, we further develop a large-scale and high-resolution CatMask-HQ annotation tailored specifically for AFHQ dataset based on analysis.

To summarize, we make the following contributions:
\begin{itemize}
\item We propose a novel framework (MaTe3D) for portrait editing. The framework can perform more effective mask-guided and text-based editing simultaneously in a single model, ensuring editing quality and stability.

\item We develop a novel 3D generator and a distillation strategy in our framework. Our generator learns global and local representations via proposed SDF and density consistency losses to enhance the mask-based editing. Our Condition Distillation on Geometry and Texture (CDGT) mitigates visual ambiguity, avoids mismatch between texture and geometry, and produces stable texture and convincing geometry while editing.

\item We create a new cat face dataset named CatMask-HQ with large scale and high-quality annotations. The dataset extends the assessment of model generalization to domains beyond human facial.

\item We conduct extensive qualitative and quantitative experiments, which demonstrate that our method can perform mask-guided and text-based portrait editing simultaneously with higher quality and stability.

\end{itemize}

\begin{figure*}[t]
  \centering
  \includegraphics[width=1.0\linewidth]{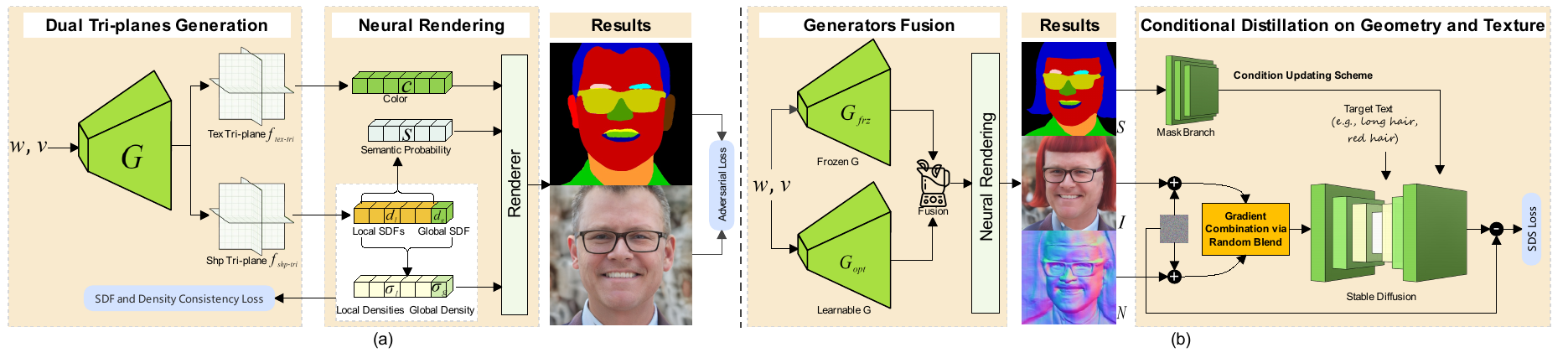}\vspace{-5pt}
  \caption{Framework of MaTe3D. MaTe3D Generator (a) consists of Tri-planes Generation and Neural Rendering. Tri-planes Generation constructs 3D volumes of texture and shape in tri-plane representations. Neural Rendering renders 3D-aware image and mask from learned SDF and color. In Inference-optimized editing phase (b), we utilize both a frozen generator ($G_{frz}$) and a learnable generator ($G_{opt}$), which were initialized by the generator in (a).  In addition, we propose Condition Distillation on Geometry and Texture (CDGT) to achieve mask-guided and text-based editing while maintaining consistent texture across multiple views and producing reasonable geometry.}\vspace{-15pt}
  \label{fig:framework}
\end{figure*}

\section{Related Works}

\subsection{3D-aware GAN based Synthesis and Editing} 

Shocked by NeRF~\cite{nerf} and SIREN~\cite{siren} in exceeding the past method with regards to multi-view image synthesis, recent methods start to focus on 3D-aware models without multi-view supervision. GRAF~\cite{graf} and pi-GAN~\cite{pigan} are pioneers in exploring NeRF-based 3D-aware synthesis and render view-consistent results. StyleNeRF~\cite{stylenerf} and CIPS-3D~\cite{cips3d} adopt progressively upsampling to synthesize high-quality 3D-aware images. StyleSDF~\cite{stylesdf}, which merges NeRF with Signed Distance Function (SDF) \cite{deepsdf}, achieves comparative performance to aforementioned methods and also produces 3D smooth complex shapes. EG3D~\cite{eg3d} introduces a hybrid explicit-implicit 3D representation that produces high-resolution image and brings implicitly learned mesh to a new level of meticulosity. However, none of the above methods consider fine-grained locally-controllable face editing with other conditions.

To achieve fine-grained level control with mask, FENeRF~\cite{fenerf} pioneers local face editing using decoupled latent codes to generate corresponding facial semantics and texture in a spatially aligned 3D volume with shared geometry. IDE-3D~\cite{ide3d} and NeRFFaceEditing~\cite{nerffaceediting} allow for semantic mask guided local control of images in real time. CNeRF~\cite{cnerf} and 3D-SSGAN~\cite{3dssgan} aim to achieve part-level editing through the compositional synthesis. For fine-grained editing with text, LENeRF~\cite{lenerf} achieves local manipulation via establishing 3D attention map via CLIP~\cite{clip}. StyleGANFusion~\cite{styleganfusion} and FaceDNeRF~\cite{facednerf} distill knowledge from diffusion model to enable manipulate Face NeRFs in zero-shot learning based on text prompts.

\subsection{Diffusion Models Guided Editing}

In recent years, diffusion models have made great strides in the field of text-to-2D generation. Some works have begun to use diffusion model priors to edit 3D scenes through text guidance. For example, Instruct-NeRF2NeRF~\cite{haque2023instruct} employs Instruct-Pix2Pix~\cite{instructp2p} to modify NeRF content using language instructions. The approach optimizes both the underlying scene and iterative optimization, resulting in a 3D scene that conforms to the input text. Blended-NeRF~\cite{gordon2023blended} uses a pre-trained language image model to synthesize objects based on textual prompts and blend them into existing scenes using 3D multi-layer perceptron (MLP) models and volume blending techniques for natural appearance. 

Score Distillation Sampling (SDS) from DreamFusion~\cite{dreamfusion} provides a new way to apply diffusion models to 3D scenes. DreamEditor~\cite{zhuang2023dreameditor} allows controllable editing of neural fields using textual prompts through SDS. FocalDreamer~\cite{li2023focaldreamer} merges shapes with editable parts based on SDS for fine-grained editing in desired regions. SKED~\cite{mikaeili2023sked}, which edits 3D shapes represented by NeRFs, utilizes two guiding sketches without altering existing fields much, supporting object creation or modification within designated sketch regions. To improve SDS, HeadSculpt~\cite{headsculpt} introduces prior-driven score distillation with landmark-guided ControlNet to improve 3D-head generation results. HeadArtist~\cite{headartist} develops self-score distillation for generating geometry and texture, leading to a significant performance boost.

\section{Preliminaries}
\label{sec:preliminary}

\subsection{ControlNet}

ControlNet~\cite{controlnet} is an innovative approach that enhances Stable Diffusion by incorporating additional conditions(canny edge, pose maps, etc.). In MaTe3D, we train two ControlNets on color mask for face and cat, respectively. The mask representation not only supports interactive editing but also enables inpainting pipeline. To achieve this, we employ a mask encoder to map the condition (e.g., mask) to the same dimensions as the latent code in Stable Diffusion. The optimization process can be expressed as follows:
\begin{align}
\mathcal{L}=E_{\epsilon \sim N(0,I),t}||\epsilon-\epsilon_{\phi}(z_t,y,S,t)||^2,
\end{align}
where $z_0$ represents the latent feature map encoded from training data. The variables $y$ and $S$ represent the text condition and mask condition, respectively. The variable $t$ represents the timestep. Additionally, $\epsilon$ is an additive Gaussian noise term. $\epsilon_{\phi}$ represents the parameters of learnable mask encoder.

The noisy data at timestep $t$ is denoted as $z_t$ and $z_t=\alpha_t z_0+\sigma _t \epsilon$, where $\alpha_t$ and $\sigma _t$ are predefined functions decided by timestep $t$.

\subsection{Score Distillation Sampling (SDS)}
Score Distillation Sampling (SDS) is developed by DreamFusion~\cite{dreamfusion} to distill knowledge from a pre-trained diffusion model into a differentiable 3D representation generator, such as NeRF~\cite{nerf} or DMTet~\cite{dmtet}. Consider a 3D representation generator $g$ is parameterized by $\theta$, we can obtain the rendered image $I=g(\theta)$. Then, given the noisy image $z_t$, text embedding $y$ and noise timestep $t$, a random noise $\epsilon$ is added to latent code $z$ encoded from image $I$. The pre-trained diffusion model $\epsilon_{\phi}$ is then adopted to predict the added noise from the noisy image. The SDS provides gradient to update the generator and can be represented as follows:

\begin{align}
\nabla_{\theta} \mathcal{L}_{SDS}(\phi, \theta)=\mathbb{E}_{\epsilon,t} [w(t)(\epsilon-\epsilon_{\phi}(z_t,y,t) \frac{\partial z}{\partial I}\frac{\partial I}{\partial \theta}],
\end{align}
where $\epsilon \sim \mathcal{N}(0, I)$, $t \sim \mathcal{U}(0.02, 0.98)$.

\section{Methods}

\minisection{Problem setting.} Our goal is to perform mask-guided and text-based portrait editing in a single model, enabling high quality and stability of 3D editing. We decouple MaTe3D into two steps. First, we propose a new 3D generator based on SDF to improve the mask-based editing performance in local areas. Our approach uses both SDF and density consistency losses to learn semantic mask fields by joint modeling of the global and local representations. Second, considering when directly manipulating the generator with both mask and text information, the system suffers from: \textit{unstable texture}  and \textit{unconvincing geometry}. Hence, we further introduce Condition Distillation on Geometry and Texture (CDGT).

\minisection{Overview.} We outline the proposed \textit{MaTe3D} pipeline in Fig.~\ref{fig:framework}, which consists of two key parts: the proposed generator in the first step (Fig.~\ref{fig:framework}(a)) and the inference-optimized editing (Fig.~\ref{fig:framework}(b)) in the second step. In Sec.~\ref{sec:sdf_generator}, we provide more details of the proposed generator, including the generator architecture, the SDF-based neural rendering  and the training objectives. In  Sec.~\ref{sec:editing}, using the well-trained generator from the first step, we introduce to fuse a frozen generator and a learnable generator, and a  Conditional Distillation on Geometry and Texture (CDGT) method to update the fused generator.

\subsection{MaTe3D Generator}
\label{sec:sdf_generator}
 
As shown in Fig.~\ref{fig:framework}(a), our MaTe3D generator contains  three parts: {1) \textit{Dual tri-planes generation} for enhancing specific facial details to capture both global and local representations, 2) \textit{Neural rendering} to incorporate both color and semantic features into the renderings, and 3) \textit{Optimization} to learn global and local representations simultaneously. In the following paragraphs, we present detailed descriptions for each part.  

\subsubsection{Dual Tri-planes Generation} 

Let  $v$ and $w$ be a camera pose and a latent code, respectively. 
Dual Tri-planes Generation takes camera pose $v$ and latent code $w$ as inputs, and generates texture tri-plane $f_{tex-tri}$ and shape tri-plane $f_{shp-tri}$. 
The dual-generation process allows for the enhancement of specific facial details, such as glasses, as shown in Fig.~\ref{fig:ablation_consistency}. The texture feature $f_{tex-tri}$ and shape feature $f_{shp-tri}$ are extracted by projecting coordinates onto the tri-planes. Subsequently, two separate MLP decoders are adopted to learn color features $c$ and Signed Distance Function (SDF) values $d=\{d_g,d_{l1}, d_{l2}, ..., d_{ln}\}$. The SDF values can not only capture the global facial structure $d_g$ but also the local components like the nose, eyes, and mouth, represented as $\{d_{l1}, d_{l2}, ..., d_{ln}\}$, where $n$ denotes the number of semantic parts.

\subsubsection{Neural Rendering} 
As shown in Fig.~\ref{fig:framework}(a), in Neural Rendering we utilize SDF values to generate density values through \textit{Density feature generation}. Next, we render a semantic map from local SDF values using \textit{Implicit semantic generation} to enable mask-based editing. Finally, by combining density, color, and semantic probabilities, we propose \textit{Feature map generation} to produce high-resolution portrait $I^h$ and semantic mask $S^h$.

\paragraph{\textit{Density Feature Generation}} Inspired by both StyleSDF~\cite{stylesdf} and VolSDF~\cite{volsdf}, we transform the signed distance values into density values for volume rendering. This transformation is formulated as follows:
\begin{align}
&\sigma_g=K_{\alpha}(d_g)=\frac{1}{\alpha}\cdot \textit{Sigmoid} (\frac{-d_g}{\alpha}),\\
&\sigma_{li}=K_{\alpha}(d_{li})=\frac{1}{\alpha}\cdot \textit{Sigmoid} (\frac{-d_{li}}{\alpha}),
\end{align}
where $\sigma_g$ is the density value of the global face,  and the $\sigma_{li}$ ($i = 1,...,n$) refer to local facial components.  The learnable  hyper-parameter $\alpha$  controls the density value tightness around the surface boundary~\cite{stylesdf}. $\alpha$ is shared among both global SDF values and the decomposed local SDF values.

\paragraph{\textit{Implicit Semantic Generation}}  
To support mask-based editing, \textit{Implicit semantic generation} is proposed to render the semantic mask. Different from conventional methods that predict a 3D semantic field, our approach leverages the correlation between semantic information and decompositional geometries. By directly converting SDF into a semantic representation within a generative radiance field, we maintain semantic consistency within geometric classes and prevent abrupt changes across different classes.

Inspired by scene representation in~\cite{objectsdf}, we directly convert SDF into semantic probabilities in generative radiance field:
\begin{align}
s_i = \frac{\gamma}{1+exp(\gamma d_i)},
\end{align}
where $\gamma$ is a hyper-parameter to control the smoothness of the function.

\paragraph{\textit{Feature Map Generation}} 
Given the previously learned density, semantic probabilities and color features, the neural renderer generates both a color feature map $C_{feat}$ and a low-resolution semantic map $S^{thumb}$. The first $3$ channels of $C_{feat}$ store the low-resolution portrait $I^{thumb}$. These maps are used to synthesize a high-resolution portrait $I^{h}$ and semantic mask $S^{h}$ through a dual StyleGAN-based sampler, which is modulated by corresponding latent codes.

\subsubsection{Optimization} 

Optimizing a SDF-based generator is challenging when learning the global and the local representations simultaneously. Thus, we propose both the SDF and the density consistency losses. The full loss function consists of several parts which are listed as follows:

\paragraph{\textit{SDF Consistency Loss}} 
In MaTe3D generator, we adopt the global SDF that defines the overall shape of the face and a set of local SDFs that capture the local parts of the face. To ensure geometrical consistency across the whole model, it is essential to integrate these local SDFs with the global SDF. This integration guarantees that all local shapes blend seamlessly, resulting in an accurate 3D representation and further rendering anti-aliasing masks. In this study, we achieve the integration by computing the loss between the minimum value of the local SDFs and the global SDF. The loss can be formulated as follows:

\begin{align}
\mathcal{L}_{sdf}=||d_{g},\min(d_{l1},d_{l2},...,d_{ln})||^2,
\end{align}

\paragraph{\textit{Density Consistency Loss}} 

In addition to the geometrical consistency achieved through the SDF consistency loss, we propose a density consistency loss to enhance the representation of the 3D model. The density consistency loss is computed by computing the difference between the local and global density values from learned signed distance values. For quantifying discrepancies in density across the entire object volume, the density consistency loss guarantees a stable training, and a better image quality. The loss can be formulated as follows:

\begin{align}
\mathcal{L}_{\sigma}=||\sigma_{g},\sum_{i=1}^{n}{\sigma_{li}}||^2,
\end{align}

\paragraph{\textit{Dual Adversarial Loss}}
We use a dual discriminator to model the distribution of portrait and semantic mask. The discriminator takes concatenation of the portrait image $(I^{thumb},I^{h})$ and semantic mask $(S^{thumb},S^{h})$ as input. We apply classic non-saturating adversarial loss with R1 regularization to obtain $\mathcal{L}_{adv}$.

\paragraph{\textit{Regularization Losses}}
We introduce three regularization losses: Eikonal loss ($\mathcal{L}_{eik}$) to ensure the physical valid of SDFs~\cite{stylesdf}, minimal surface loss ($\mathcal{L}_{surf}$) to prevent the SDFs from modeling spurious and non-visible surfaces, and density regularization loss ($\mathcal{L}_{reg}$) to regularize the density convert from SDF to prevent "seam" artifacts. 

In summary, the final loss of SDF-based generator is:
\begin{equation}
\begin{aligned}
    \mathcal{L}_{g}&=\lambda_{adv}\mathcal{L}_{adv} + \lambda_{sdf}\mathcal{L}_{sdf}+ \lambda_{\sigma}\mathcal{L}_{\sigma} \\&+\lambda_{eik}\mathcal{L}_{eik}+\lambda_{surf}\mathcal{L}_{surf}+\lambda_{reg}\mathcal{L}_{reg},
\end{aligned}
\end{equation}
where the hyper-parameters $\lambda_{adv}$, $\lambda_{sdf}$, $\lambda_{\sigma}$, $\lambda_{eik}$,$\lambda_{surf}$ and $\lambda_{reg}$ balance the contribution of each loss term, respectively.

\begin{figure}[!t]
  \centering
  \includegraphics[width=1.0\linewidth]{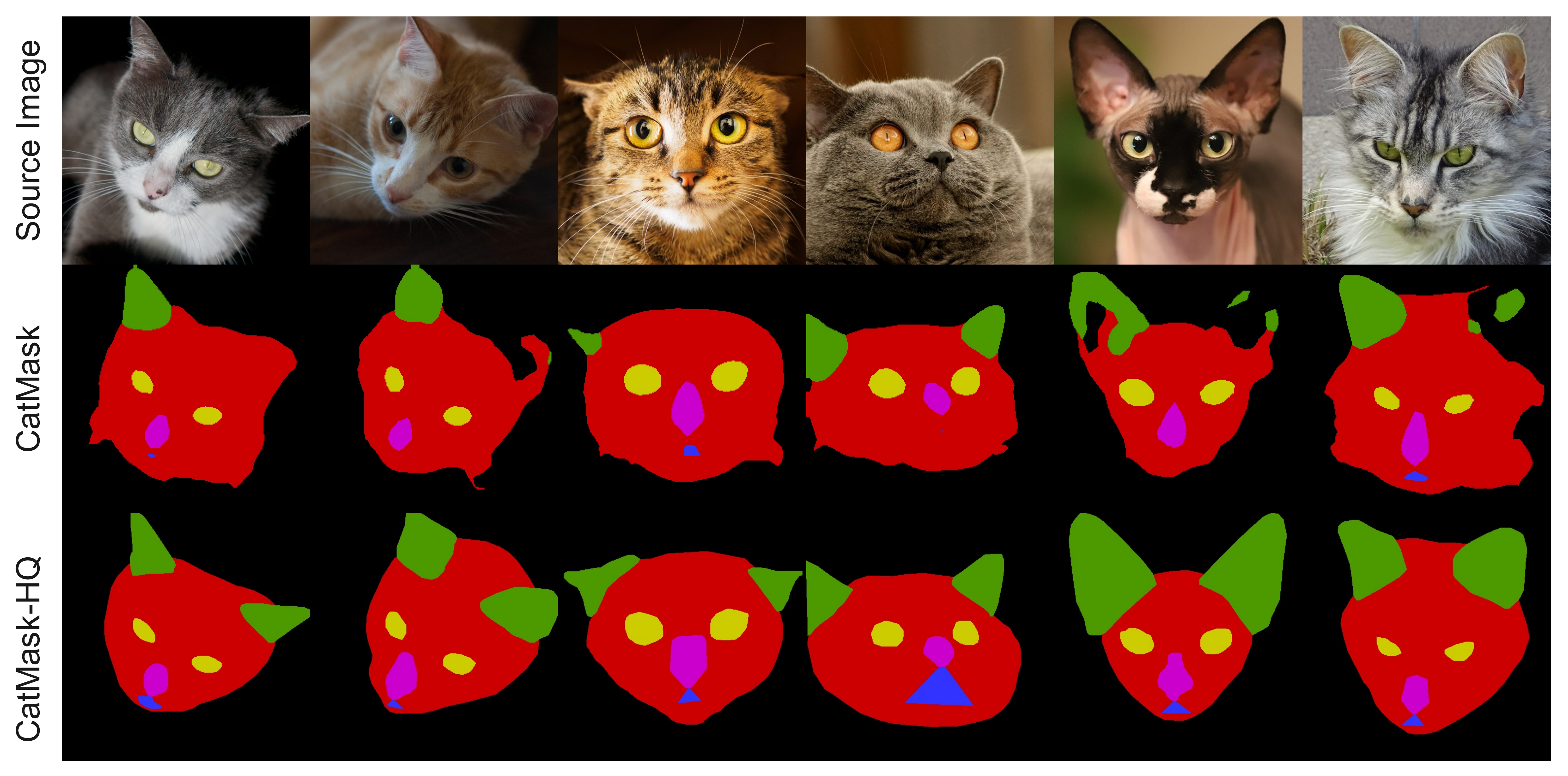}
  \caption{Dataset qualitative comparisons with an existing dataset. CatMask-HQ has superior annotation quality.}
  \label{fig:cat_data}
\end{figure}
\begin{figure*}[t]
  \centering
  \includegraphics[width=1.0\linewidth]{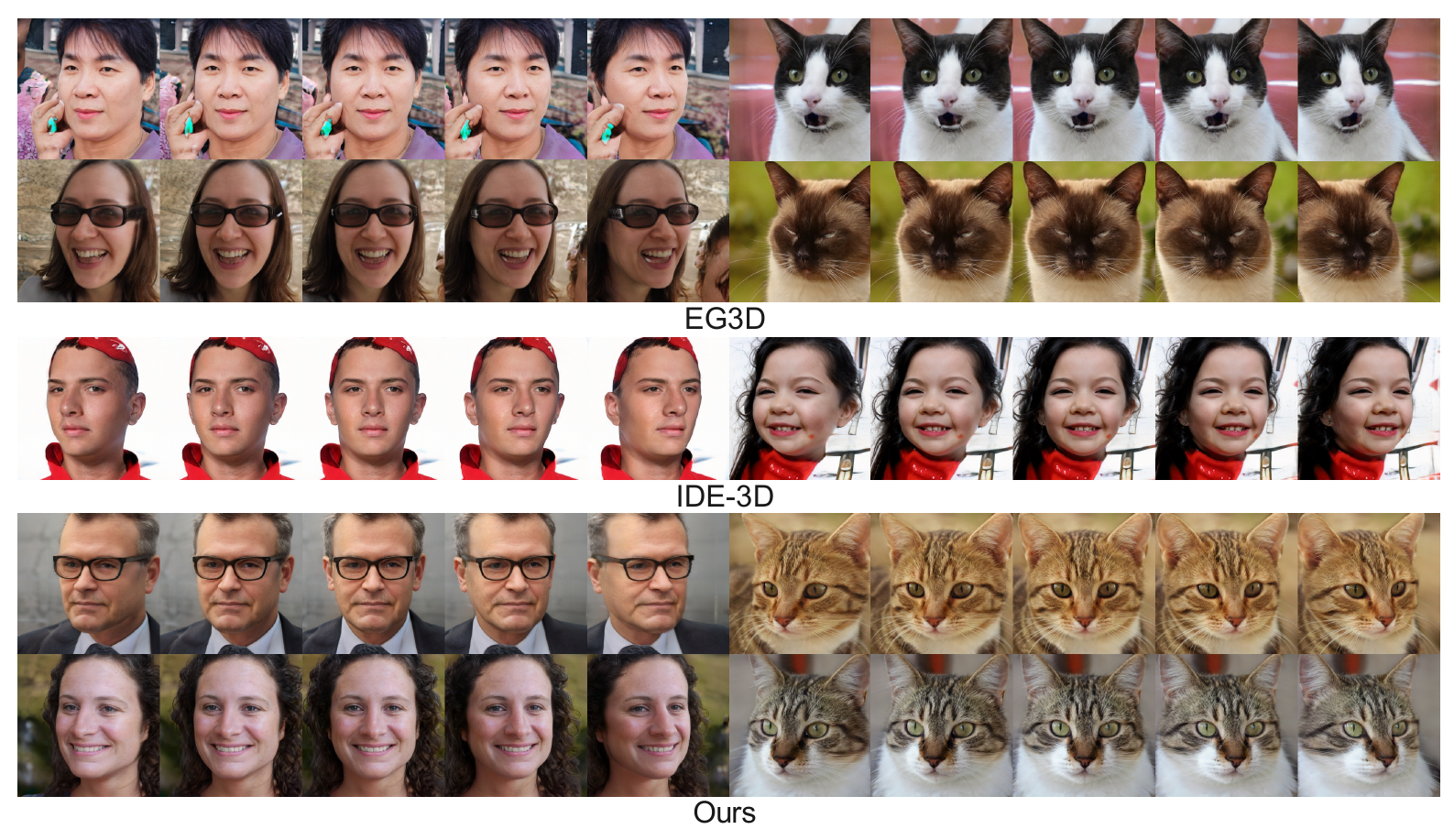}
  \caption{Qualitative comparison in image quality. MaTe3D achieves comparable image quality and better view consistency across face and cat.}
  \label{fig:quality}
\end{figure*}

\subsection{Inference-optimized Editing}
\label{sec:editing}
The Inference-optimized Editing aims to modify the content and structure of the input image $I$ using a customized semantic mask $S^{target}$ and a text prompt $y$ (e.g. 'a woman with wrinkles on face'). As shown in Fig.~\ref{fig:framework}(b), it includes \textit{Generators Fusion} and \textit{Conditional Distillation on Geometry and Texture}. In \textit{Generators Fusion} module, we present a frozen and a learnable generators, respectively. Both of them are used to generate 3D masks for feature fusion. In \textit{Conditional Distillation on Geometry and Texture}, we aim to  address both distortion-free textures and convincing geometries when performing both the mask-guided and the text-based 3D-aware portrait editing.

\minisection{Generator Fusion}
Generator Fusion uses the camera pose $v$ and latent code $w$ to generate pairs of texture tri-planes and shape tri-planes. It then fuses the tri-planes by incorporating 3D masks derived from semantic probabilities.
In Fig.~\ref{fig:framework}(b), we illustrate the Generator Fusion of inference-optimized editing phase in MaTe3D, which comprises a frozen generator $G_{frz}$ and a learnable generator $G_{opt}$. Both $G_{frz}$ and $G_{opt}$ are initialized based on the well-trained generator from Sec.~\ref{sec:sdf_generator}. We utilize $G_{frz}$ and $G_{opt}$ to derive 3D masks that facilitate feature fusion across tri-planes from both generators, allowing us to synthesize edited images $I$.

\minisection{Conditional Distillation on Geometry and Texture} We innovatively propose CDGT for precise and interactive 3D manipulation by leveraging a user-provided mask and textual description. By iteratively refining masks through a progressive condition updating scheme with rendered results from the generators, CDGT addresses the challenge of 3D mask annotation scarcity while enhancing texture stability while 3D editing. This mitigates visual ambiguity across diverse views within the diffusion model. Additionally, by combining gradients from both image and normal map, we distill controllable diffusion priors over texture and geometry to preserves underlying geometry integrity, avoiding mismatch between texture and geometry.

\subsubsection{Condition Updating Scheme}

The original implementation of SDS loss primarily focuses on texture, lacking the ability to control shape~\cite{headsculpt}. Some methods enhance diffusion by incorporating landmarks obtained from an off-shelf detector for better control. However, these approaches still face challenges with unstable texture due to sparse conditions and limited 3D consistency. To tackle this issue, we introduce Condition Updating Scheme. This technology regards the generated mask $m$ as a condition for ControlNet, establishing a connection between the generator and diffusion prior. We iteratively update the condition of ControlNet during editing and integrate it into multi-view consistency to ensure stable texture. This design progressively integrates diffusion priors conditioned by 3D-aware information into the 3D content. Despite initially computing incorrect gradients during optimization, the interplay between front-view supervision and Condition Updating Scheme guarantees that results converge to maintain texture consistency and align with the modified mask.

\subsubsection{Gradient Combination}

While editing a neural radiance field, maintaining convincing geometry deformation is as essential as changing texture. However, we observe that the quality of geometry degrades when editing certain complex attributes (e.g., hat in Fig.~\ref{fig:ablation_ckgt}). This degradation is attributed to a mismatch between texture and geometry caused by the absence of 3D supervision. To address this issue, we align geometry and texture by computing gradients on both the image $I$ and normal map $N$. In the approach, we introduce a novel random blending strategy to combine gradients. Specifically, we encode the image $I$ and normal map $N$ to obtain their latent representations $z^I$ and $z^N$. We then use a sampled value $\mu$ to blend texture and geometry: $\hat{z} = \mu z^{I} + (1-\mu) z^{N}$. In contrast that TADA~\cite{tada} fuses texture and geometry using a constant value, our method employs randomly sampled values $\mu$ ranging from $0$ to $1$. By incorporating randomness into the fusion process, random blending enhances the robustness of editing process by preventing overfitting on specific patterns and improving its capability to handle discrete attributes.

Given a pre-trained SDF-based generator $g$ with parameters $\theta$, we generate images $I$, normal maps $N$ and semantic masks $S$ using $(I,N,S)=g(\theta)$. Given target prompt $y$, we compute CDGT as follows:
\begin{equation}
\begin{aligned}
 \nabla_{\theta} \mathcal{L}_{CDGT}&=  \mathbb{E}_{\epsilon,t} [w(t)(\epsilon-\epsilon_{\phi}(\hat{z_t},y,S,t) \frac{\partial \hat{z}}{\partial N}\frac{\partial N}{\partial \theta} \\
& + w(t)(\epsilon-\epsilon_{\phi}(z_t^{I},y,S,t) \frac{\partial z^{I}}{\partial I}\frac{\partial I}{\partial \theta}],
\end{aligned}
\end{equation}

We iteratively optimize the learnable $G_{opt}$ using CDGT loss, ID loss $\mathcal{L}_{ID}(I,I_{origin})$, and segmentation loss $\mathcal{L}_{seg}(S_{front},S_{target})$. The weights for these loss terms are empirically defined, respectively. Here, $I_{origin}$ is generated from the frozen generator $G_{frz}$, and $S_{front}$ is rendered in the same view (front view) as $S_{target}$.

In summary, we first pretrain the proposed generators on training dataset. Then, we train ControlNet on Stable Diffusion $1.5$ with mask condition, utilizing images and annotations. Finally, when editing image, we fix both the generators and ControlNet and update a new well-initialized generator with the proposed Condition Distillation on Geometry and Texture (CDGT) loss, ID loss and segmentation loss.

\section{Experiments}

\subsection{Implementation Details}

\subsubsection{Setting of Generator Training}

We train our generator on FFHQ~\cite{stylegan} and proposed CatMask-HQ.
We adopt sphere initialization from StyleSDF~\cite{stylesdf}. The training is performed on $4$ NVIDIA A100 GPUs ($80$G) using batch size of $8$ for $50,000$ steps. We use Adam optimizer~\cite{adam} and set the learning rates for the generator and dual discriminator to $0.0025$ and $0.002$, respectively. For losses, we assign $\lambda_{adv}=1$, $\lambda_{sdf}=0.1$, $\lambda_{\sigma}=0.001$, $\lambda_{eik}=0.05$, $\lambda_{surf}=0.1$, and $\lambda_{reg}=1$. The weights  $\lambda_{adv}$, $\lambda_{eik}$, $\lambda_{surf}$ and $\lambda_{reg}$ are defined the same as those in StyleSDF~\cite{stylesdf}. We define both $\lambda_{sdf}$ and $\lambda_{\sigma}$  by tentative experiments.

\begin{figure}[!htbp]
  \centering
  \includegraphics[width=1.0\linewidth]{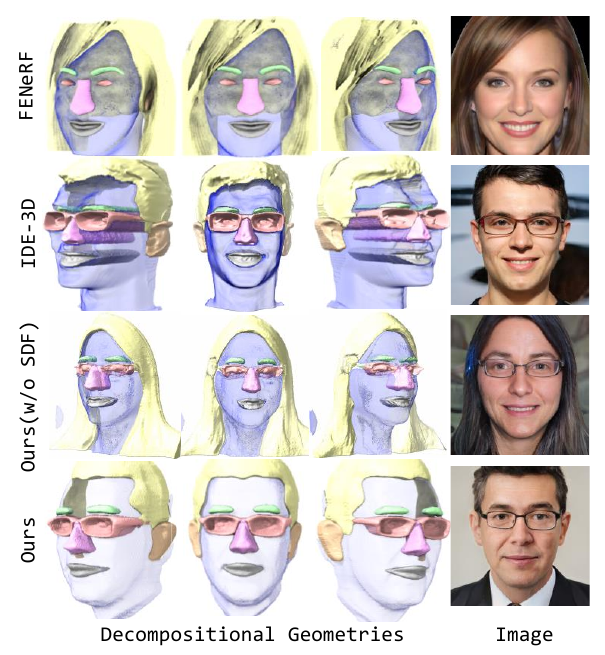}
  \caption{Qualitative comparison with the state-of-the-art methods on decompositional geometries. MaTe3D has more consistent and well-decompositional meshes than FENeRF (staircase artifacts), IDE-3D (inaccurate density distribution along rays) and Ours(w/o SDF, inaccurate eye modeling). We merge some classes for demonstration purposes.}
  \label{fig:dg}
\end{figure}


\subsubsection{Setting of Editing}

During editing phase, we optimize learnable generator $2,000$ iterations for each target mask and input text prompt, which  takes around $20$ minutes on a single NVIDIA A100 GPU ($80$G). We use Adam~\cite{adam} as the optimizer with learning rate of $3e-3$. We determine the values of $\lambda_{CDGT}$, $\lambda_{ID}$, and $\lambda_{seg}$ as $0.001$, $0.5$, and $1$ through preliminary experiments.


\begin{table}
\centering
\setlength{\tabcolsep}{0.5em}
\begin{tabular}{l|lll|ll} 
\toprule
\multirow{2}{*}{Method} & \multicolumn{3}{c|}{FFHQ}                                                   & \multicolumn{2}{c}{AFHQ}                           \\
                        & \multicolumn{1}{c}{FID$\downarrow$} & \multicolumn{1}{c}{KID$\downarrow$} & \multicolumn{1}{c|}{ID$\uparrow$} & \multicolumn{1}{c}{FID$\downarrow$} & \multicolumn{1}{c}{KID$\downarrow$}  \\ 
\midrule
    FENeRF~\cite{fenerf}    & 29.0 & 3.728 & 0.61 & / & / \\
    StyleNeRF~\cite{stylenerf} & 7.8  & 0.220  & 0.62 & 14.9 & 0.368 \\
    StyleSDF~\cite{stylesdf}  & 13.1 & 0.269 & 0.74 & 12.8 & 0.455 \\
    MVCGAN~\cite{mvcgan}  & 13.4 & 0.375 & 0.58 & 17.1 & 0.198 \\
    EG3D~\cite{eg3d}      & 4.7  & 0.132 & 0.77 & 2.7 & \textbf{0.041} \\
    IDE-3D~\cite{ide3d}    & \textbf{4.6}  & \textbf{0.130} & 0.76  & /&/ \\
    BallGAN~\cite{ballgan}    & 5.7 & 0.312 & 0.75  & 4.7 & 0.096 \\
    MaTe3D\\(w/o dual tri-planes)    & 5.8  & 0.389 & 0.75 & 3.9 & 0.106 \\
    MaTe3D    & 5.1  & 0.215 & \textbf{0.79} & \textbf{2.5} & 0.078 \\
\bottomrule
\end{tabular}
  \caption{Quantitative comparison with the state-of-the-art methods. The comparison results are quoted from IDE-3D~\cite{ide3d}, EG3D~\cite{eg3d} and BallGAN~\cite{ballgan}.}
  \label{tab:quality}
\end{table}



\subsection{CatMask-HQ}

Current mask-based editing methods are primarily validated only on CelebA-HQ Mask~\cite{maskgan} or FFHQ~\cite{stylegan}
While some few-shot techniques have produced faces and masks in other domains, such as cat faces~\cite{sem2nerf}, both the image and the mask quality are subpar with noticeable aliasing~\cite{kangle2023pix2pix3d}.

To expand the scope beyond human face and explore the model generalization and expansion, we design the CatMask-HQ dataset with the following representative features:

\begin{itemize}
    \item \textbf{Specialization:}  CatMask-HQ is specifically designed for cat faces, including precise annotations for six facial parts (background, skin, ears, eyes, nose, and mouth) relevant to feline features.
    \item \textbf{High-Quality Annotations:} The dataset benefits from manual annotations by $50$ annotators and undergoes $3$ accuracy checks, ensuring high-quality labels and reducing individual differences.
    \item \textbf{Substantial Dataset Scale:} With approximately $5,060$ high-quality real cat face images and corresponding annotations, CatMask-HQ provides ample training database for deep learning models.
\end{itemize}

\begin{figure*}[htbp]
  \centering
  \includegraphics[width=\linewidth]{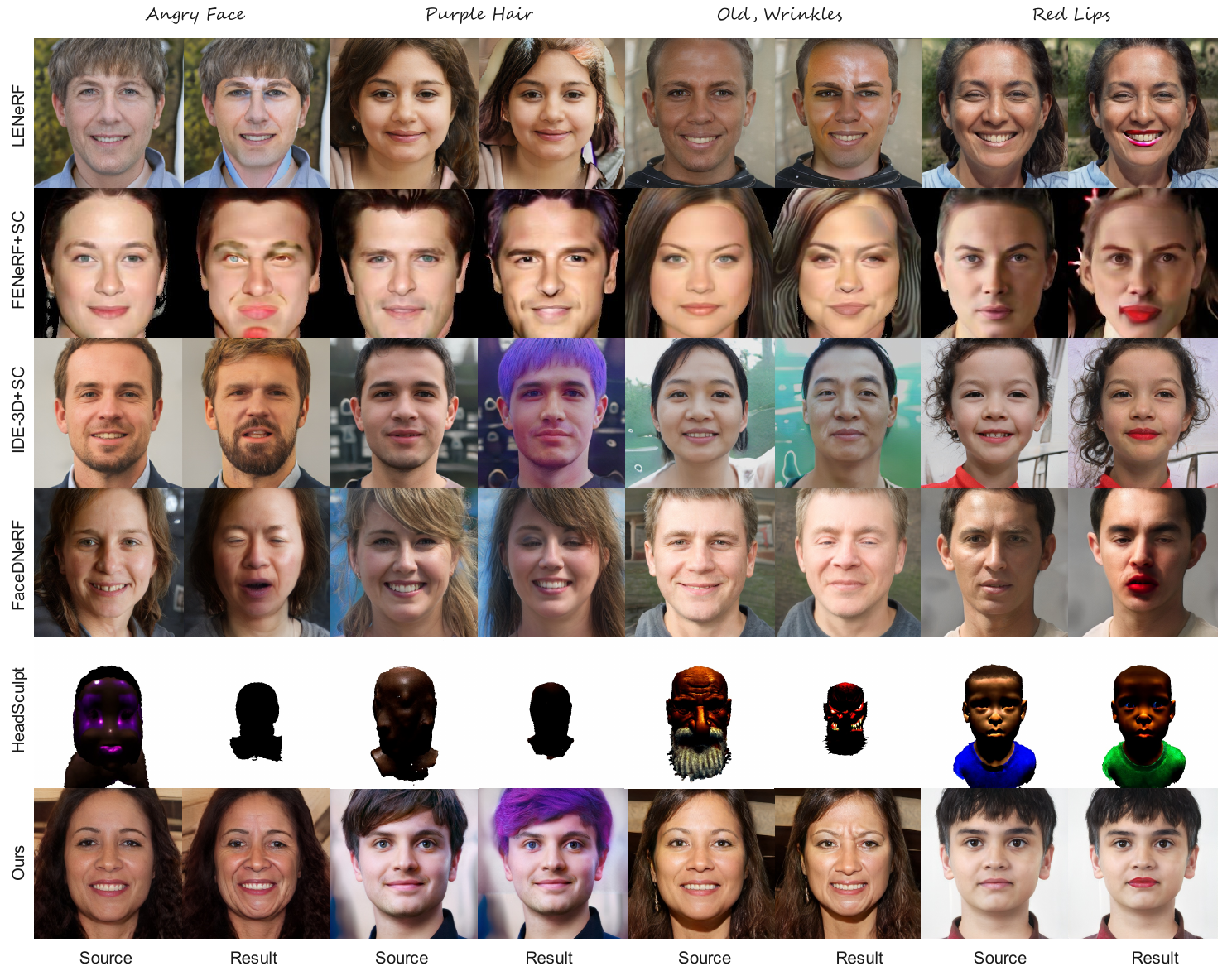}
  \caption{Text-based Editing. Our method can edit images using text while preserving the mask stability, unlike other methods that fail to maintain an unchanged mask or synthesize unimpressive texture. Here, 'SC' represents 'StyleCLIP'.}
  \label{fig:textedit}
\end{figure*}  

For the fair comparison among prior works, we utilize the same segmentation network as CatMask to label AFHQ and produce CatMask. Qualitative comparisons between the pseudo labels and annotated results are depicted in Fig.~\ref{fig:cat_data}. Our proposed dataset boasts more precise annotations devoid of aliasing edges or empty classes found in the pseudo labels.

\begin{table}[t]
  \centering
  \resizebox{1\linewidth}{!}{
  \begin{tabular}{l|l|l}
    \toprule
    Method  & Cham. $\downarrow$ & Norm.$\uparrow$\\
    \midrule
    IDE-3D~\cite{ide3d} \textit{w} nose  & 3039.3770 & 0.0443 \\
    IDE-3D~\cite{ide3d} \textit{w} nose, mouth  & 3084.0366 & 0.0445 \\
    IDE-3D~\cite{ide3d} \textit{w} nose, mouth, hair  & 2607.2324 & 0.0450 \\
    IDE-3D~\cite{ide3d} \textit{w} nose, mouth, hair, ear  & 2677.9805 & 0.0451 \\
    \midrule
    MaTe3D  \textit{w} nose & 1419.5851 & 0.0934 \\
    MaTe3D  \textit{w} nose, mouth & 1424.8981 & 0.0927 \\
    MaTe3D  \textit{w} nose, mouth, hair & 528.1699 & 0.0877 \\
    MaTe3D  \textit{w} nose, mouth, hair, ear & \textbf{528.7918} & \textbf{0.0878} \\
    \bottomrule
  \end{tabular}
  }
  \caption{\textbf{Evaluation on Chamfer-distance (L1)} and \textbf{normal consistency score (Norm.)~\cite{mescheder2019occupancy}} between representative editable 3D-aware generative model with MaTe3D.}
  \label{tab:mesh_consist}
\end{table}

\subsection{Baselines} 
We evaluate the generator of MaTe3D against state-of-the-art genertors, including FENeRF~\cite{fenerf}, StyleNeRF~\cite{stylenerf}, StyleSDF~\cite{stylesdf}, EG3D~\cite{eg3d}, MVCGAN~\cite{mvcgan}, BallGAN~\cite{ballgan} and IDE-3D~\cite{ide3d}, respectively. We evaluate MaTe3D from two aspects including text-based and mask-text-hybrid editing. For text-based editing, we compare MaTe3D with LENeRF, FENeRF+StyleCLIP, IDE-3D+StyleCLIP, FaceDNeRF~\cite{facednerf} and HeadSculpt~\cite{headsculpt}, respectively. We reimplement LENeRF~\cite{lenerf} and HeadSculpt~\cite{headsculpt} strictly according to their papers. We reproduce FENeRF+StyleCLIP and IDE-3D+StyleCLIP for comparison purposes, similarly to LENeRF~\cite{lenerf}. In hybrid editing, we compare two baselines: FENeRF+StyleCLIP and IDE-3D+StyleCLIP with mask and text editing simultaneously.

\subsection{MaTe3D generator Evaluation}

\begin{figure*}[t]
  \centering
  \includegraphics[width=\linewidth]{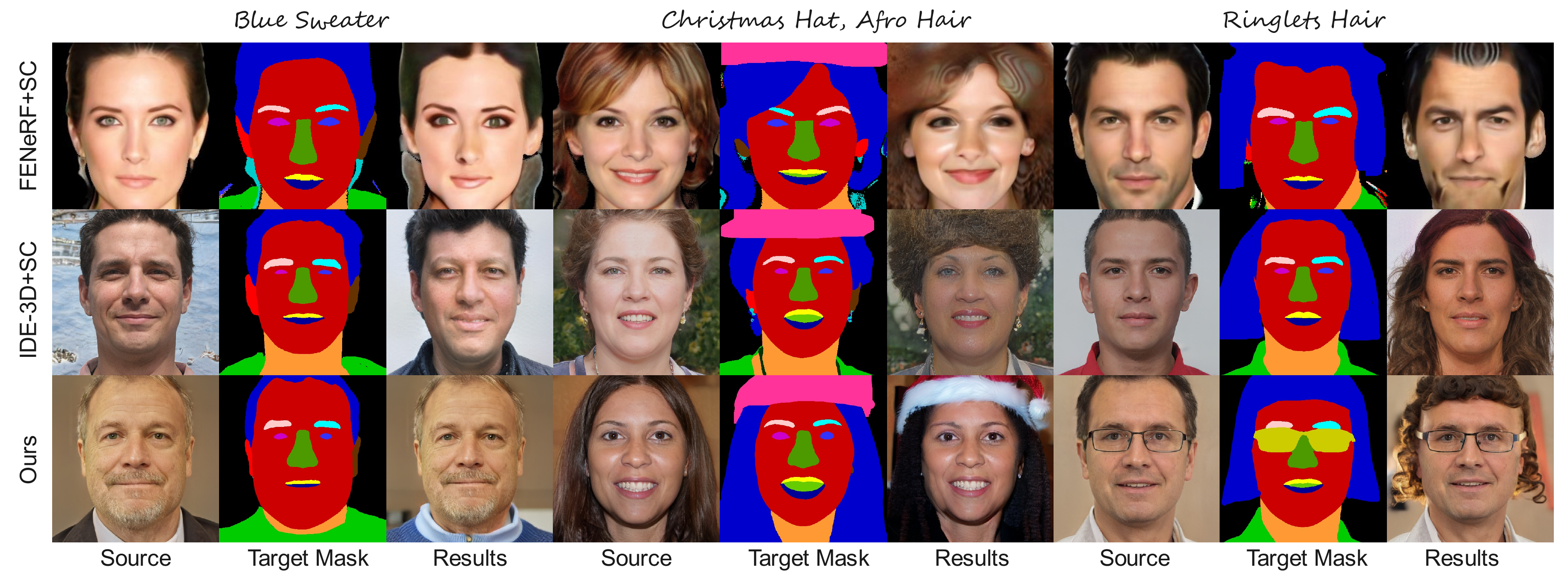}
  \caption{Mask-guided Text-based Editing. Our method allows for synchronous editing of images using both text and mask. The generated images meet the above two conditions simultaneously. Other methods fail to balance the two conditions and produce unreasonable results.}
  \label{fig:hyperedit}
\end{figure*}

\begin{figure*}[t]
  \centering
  \includegraphics[width=1.0\linewidth]{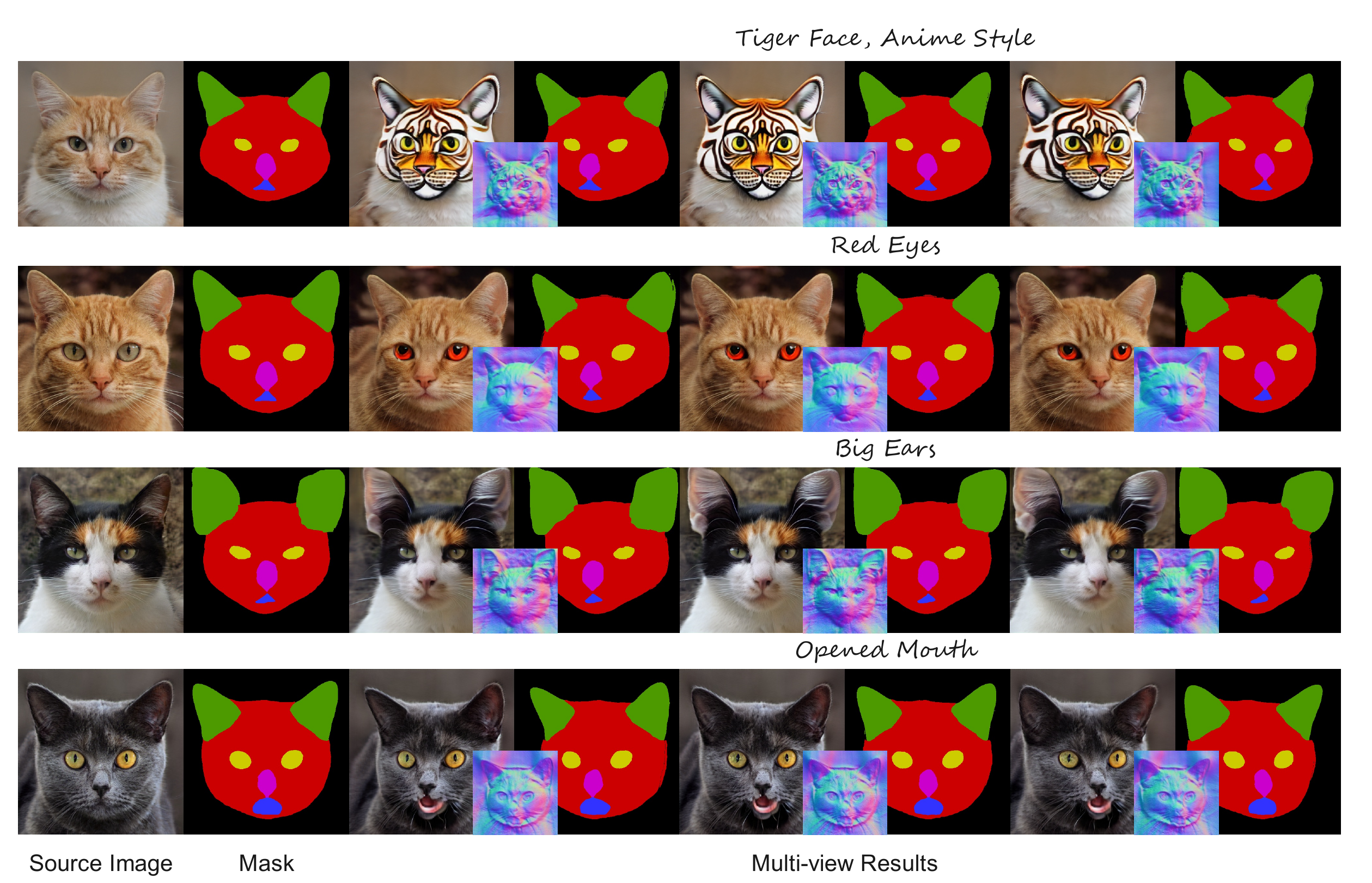}
  \caption{Text-based editing (Tiger Face, Anime Style and Red Eyes). Mask-guided text-based editing (Big Ears and Opened Mouth).}
  \label{fig:cat_edit}
\end{figure*}

\minisection{Generator Performance: qualitative and quantitative results.}

We adopt three metrics to comprehensively evaluate the synthesis quality of generated images quantitatively. Both the lower Frechet Inception Distance (FID)~\cite{fid} and Kernel Inception Distance (KID)~\cite{kid} indicate higher quality of synthesized images. Multi-view identity consistency (ID)~\cite{id} calculates the face similarity under different sampled camera pose. As shown in Tab.~\ref{tab:quality}, MaTe3D achieves comparable results with SOTAs in terms of FID, KID and ID. Fig.~\ref{fig:quality} presents a qualitative comparison of MaTe3D against 3D-aware GANs. It is evident that MaTe3D achieves superior image quality and better view consistency than the baselines, indicating the effectiveness of both the global and the local SDF representations in the generator. Also, we present additional results for CatMask-HQ, and compare the image quality with that by EG3D~\cite{eg3d}, as depicted in Fig.~\ref{fig:quality} and Tab.~\ref{tab:quality}. Our method produces better image quality with better view consistency.

Then we evaluate the geometry quality of MaTe3D, which contributes to accurate mask and provides view consistency results.

\minisection{Decompositional Geometries: qualitative result.} We compare MaTe3D with two other state-of-the-art methods: FENeRF and IDE-3D. As shown in Fig.~\ref{fig:dg},  MaTe3D outperforms its predecessors. FENeRF produces visually irritating staircasing artifacts~\cite{morsy2022shape} (the so-called bulls eye effect~\cite{IDW1968}) that are particularly noticeable when viewing small objects up close. It is possibly caused by the sparse and clustered sampling points in the frustum of FENeRF. IDE-3D  produces photorealistic semantic-aware images of relatively compact  whole faces. However,  it fails to maintain a well-established zero level-set in its neural radiance field,  resulting in constant density distribution along rays even for local parts like nose, mouth and eyes. The situation may be caused by ignoring the association between geometries and semantic information,  which leads to the mis-disentanglement of underlying local geometries.

\minisection{Decompositional Geometries: quantitative result.} We aim to investigate the compositional consistency in Mate3D in quantitative way. We combine the local facial meshes into a unified mesh, and then compare it with the original whole face.  The results, presented in Tab.~\ref{tab:mesh_consist}, demonstrate that MaTe3D significantly enhances 3D surface reconstruction quality compared to IDE-3D. Note that we do not include FENeRF in this comparison as its produced geometries have different resolutions of MaTe3D and IDE-3D, making it an unfair comparison. MaTe3D outperforms IDE-3D by large margins in terms of Chamfer-L1 (approximately 1/5 of IDE-3D) and normal consistency (approximately two times of IDE-3D), indicating its superior performance in integrity and compositionality.

\subsection{Edit Results}

\begin{table}
  \centering
  \setlength{\tabcolsep}{2em}
  \begin{tabular}{l|l|l}
    \toprule
    CLIPScore & FFHQ & AFHQ  \\
    \midrule
    LENeRF~\cite{stylesdf}  & 0.612 & 0.512  \\
    FENeRF+SC~\cite{eg3d}      & 0.729 &    / \\
    IDE-3D+SC~\cite{ide3d}    & 0.762  &    / \\
    MaTe3D    & \textbf{0.791}  & \textbf{0.732} \\
    \bottomrule
  \end{tabular}
  \caption{Quantitative comparison with the state-of-the-art editing methods. SC in table represents StyleCLIP.}
  \label{tab:clip}
\end{table}

\minisection{Comparison with Text-based Method.} In Fig.~\ref{fig:textedit}, we show the results of text-based editing.
When using CLIP for text-based editing, LENeRF (unoffical reproduction)  struggles to produce diverse images,  and suffers from insufficient expressive ability and scalability. Although FENeRF+StyleCLP is able to generate images that match the text,  other attributes have been unexpected changed. 
Like FENeRF+StyleCLP, IDE-3D+StyleCLIP suffers from a similar issue that the model tends to generate images based on text rather than the mask. Our method produces high-quality textures that are faithful to the text prompt and meanwhile maintain structural stability.

We report the CLIPScore~\cite{clipscore} values in Tab.~\ref{tab:clip}, which replies the correlation between edited images and input texts. We achieve the best ClIPscore ($0.791$) comparing with LENeRF ($0.612$), FENeRF+StyleCLIP ($0.729$), and IDE-3D+StyleCLIP ($0.762$), respectively.

\begin{figure}[t]
  \centering
  \includegraphics[width=0.97\linewidth]{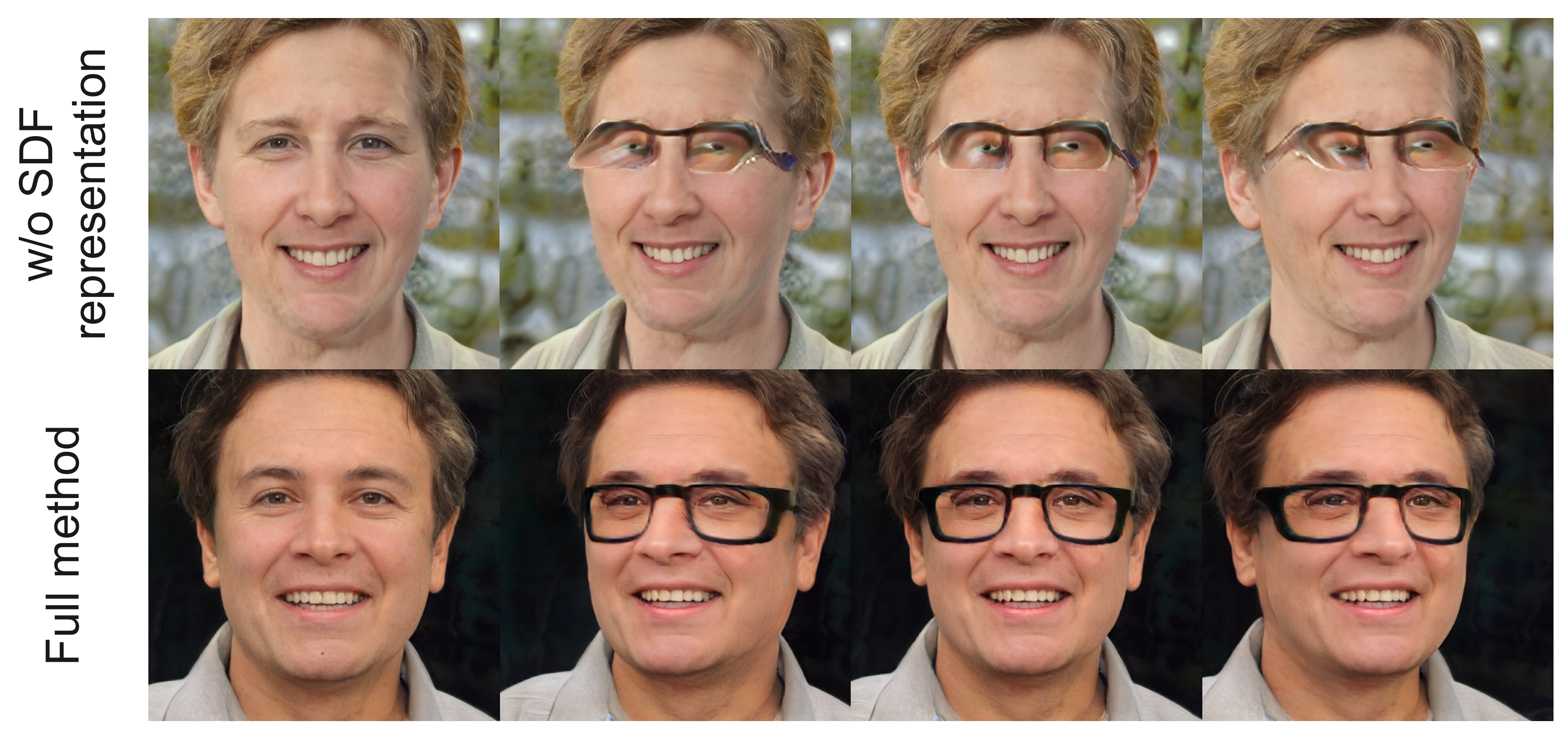}
  \caption{Ablation study of utilizing SDF representations in generator. With SDF, our method produces high-fidelity and view-consistency results.}
  \label{fig:ablation_sdf}
\end{figure}
\begin{figure}[t]
  \centering
  \includegraphics[width=0.97\linewidth]{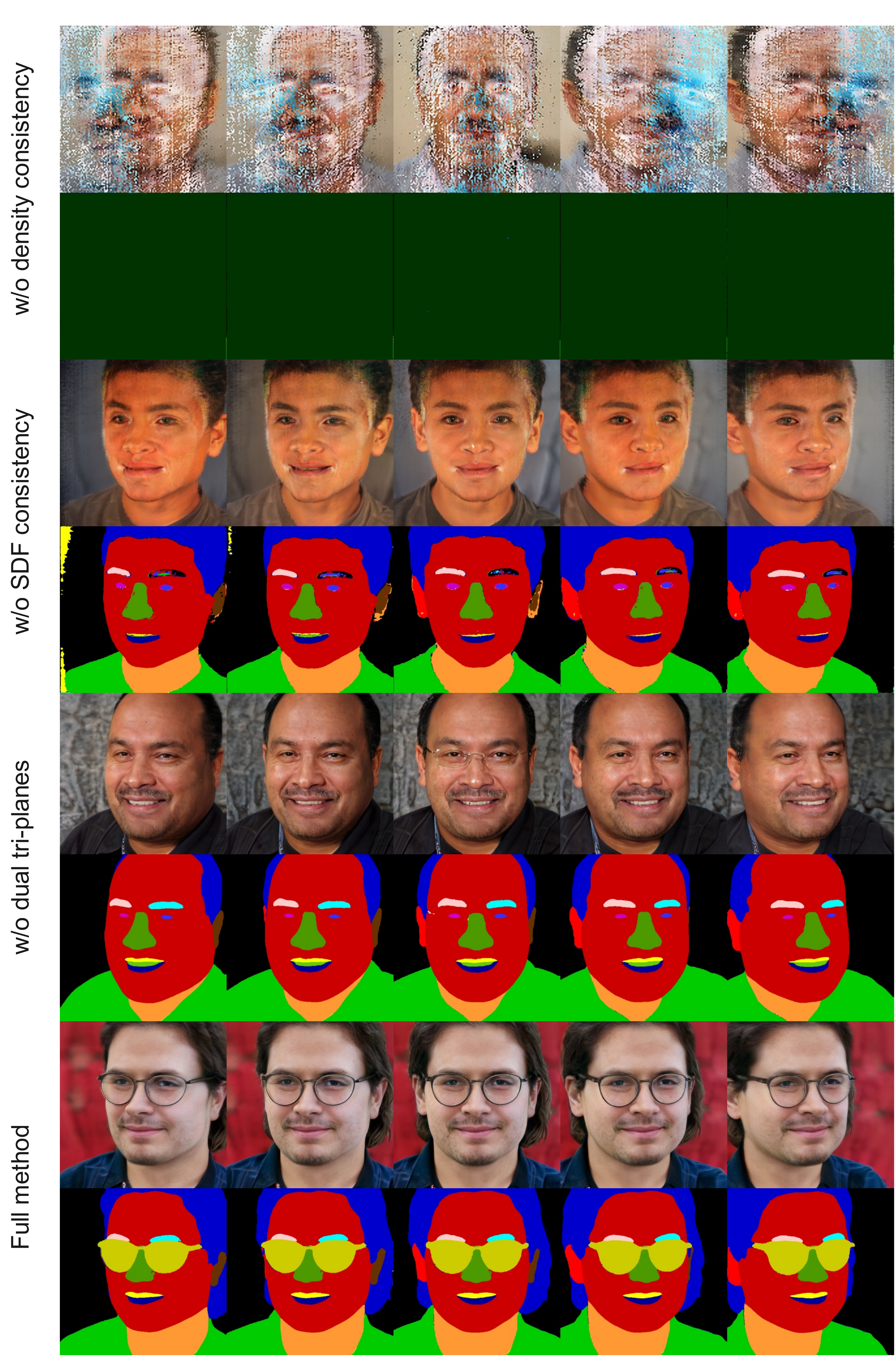}
  \caption{Ablation study for evaluating the effectiveness of utilizing density consistency loss, SDF consistency loss, and dual tri-planes generation in a generator. The generator cannot properly learn image and semantic mask without density consistency loss. Without SDF consistency loss, local representation learning impacts global representation performance causing the generator to synthesize blurry images and semantic masks with aliasing. With dual tri-plane generation, the generator can synthesize more convincing multi-view consistent results on specific details such as glasses in discrete attributes.
  }
  \label{fig:ablation_consistency}
\end{figure}
\begin{figure}[t]
  \centering
  \includegraphics[width=1\linewidth]{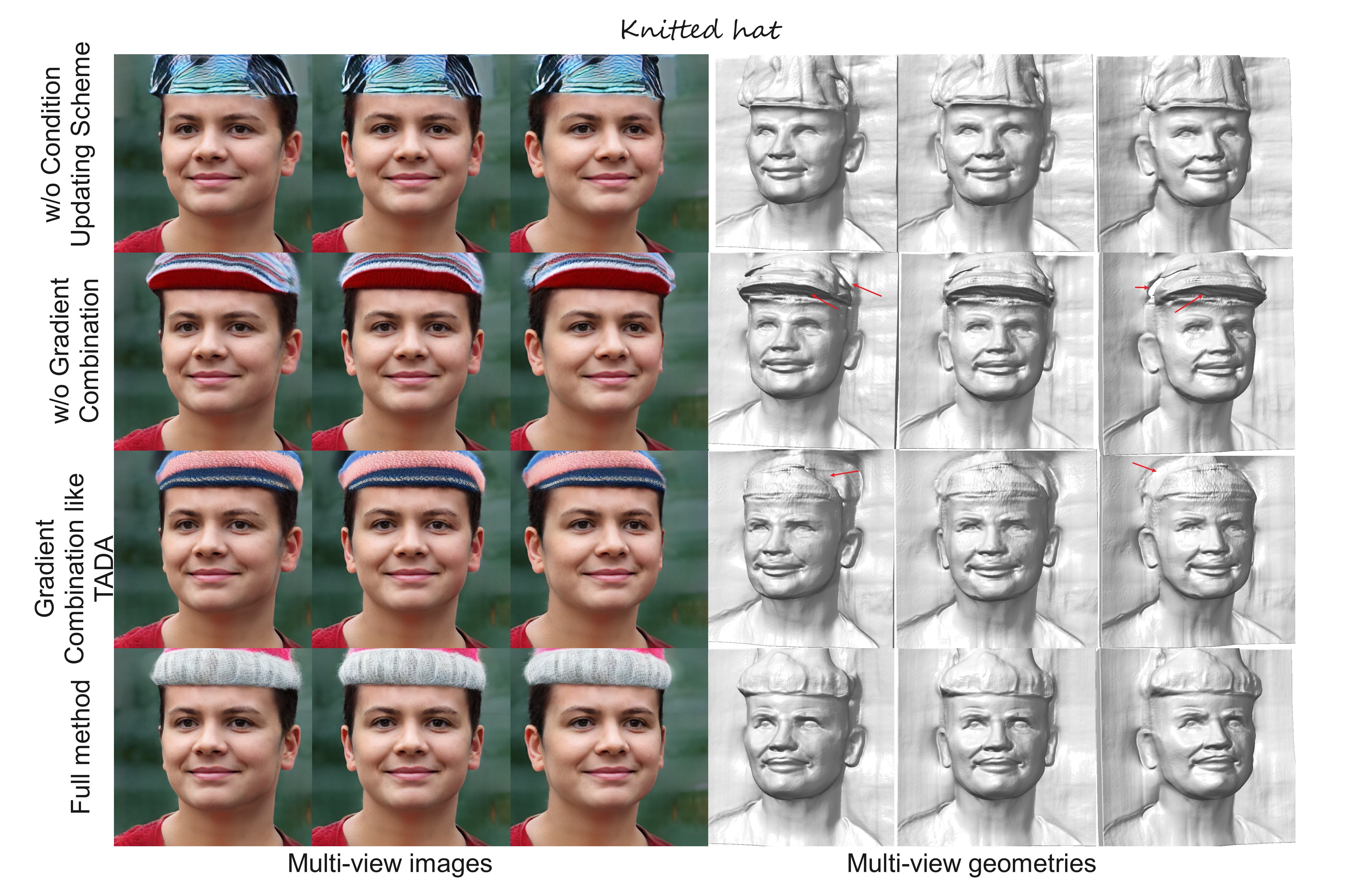}
  \caption{Ablation study on the utilization of Condition Distillation on Geometry and Texture (CDGT). The absence of Condition Updating Scheme in CDGT results in unstable texture and noticeable blurring during editing. Without combining gradients or naively mixing them with the TADA~\cite{tada} strategy, the geometry remains unconvincing and hollow, leading to a degradation in 3D quality. Our method demonstrates stable textures and convincing geometry.}
  \label{fig:ablation_ckgt}
\end{figure}

\minisection{Comparison with Mask-guided Text-based Method.}
As shown in Fig.~\ref{fig:hyperedit}, we present qualitative comparisons of our method with existing methods for both mask and text. Our results show high-quality performance in both the mask and text. However, all baselines struggle to effectively utilize both information of mask and text. Some methods prioritize preserving the mask at the expense of ignoring the text (e.g., FENeRF+StyleCLIP with prompt 'blue sweater'), while others focus solely on matching the text and disregard the mask (e.g., IDE-3D+StyleCLIP with prompt 'Christmas hat, Afro hair'). This highlights a lack of balance between the roles of mask and text during synchronous editing within baseline methods.

\minisection{Editing Results on CatMask-HQ.}
We provide more editing results in Fig.~\ref{fig:cat_edit}. Our method can synthesize stable and anti-aliasing textures, as well as maintain view consistency by synthesizing anti-degradation geometries.

\subsection{Ablation Study}

\minisection{Local SDF Representations.}
We conduct an ablation study on the local SDF representation. As depicted in Fig.~\ref{fig:dg}, Our model (w/o SDFs) generates photorealistic images and semantic-aware masks,  but suffers from unsatisfactory geometries. 
Benefiting from our generator architecture, the results without SDFs have superior geometries compared to IDE-3D~\cite{ide3d}, but errors still exist in modeling eyeglasses (incomplete edges) and nose (longer and deeper than normal). These errors occur due to a loss of connection between decompositional facial components. The poor geometry also leads to inaccurate masks.

We randomly sample $1,000$ images and masks, then use a pre-trained face segmentation model to predict the ground-truth of masks. We find that using both  global and local SDF representations can achieve a higher mIoU of $0.873$, while 'w/o SDF' only reaches $0.812$. Our method significantly improves mask accuracy with the global and local SDFs and their corresponding geometric constraints. Additionally, we observe that using global and local SDF representations helps maintain view-consistency during the editing phase (Fig.~\ref{fig:ablation_sdf}). Meanwhile, ignoring global and local representations  causes inconsistent results, particularly when adding facial accessories during editing.

\minisection{Consistency Loss and Network Architecture.} As depicted in Fig.~\ref{fig:ablation_consistency}, we perform an ablation study on the proposed consistency loss and network architecture. Our findings suggest that omitting the density consistency loss leads to a significant degradation in image quality, severely compromising the overall fidelity of the whole image. Additionally, omitting the SDF consistency loss results in aliasing edges in the semantic mask and requires more iterations to converge. Incorporating dual tri-plane generation improves image quality (see Tab.~\ref{tab:quality}) and provides more details in specific areas, such as glasses in face (see Fig.~\ref{fig:ablation_consistency}).

\minisection{Conditional Distillation on Geometry and Texture.} As shown in Fig.~\ref{fig:ablation_ckgt}, we conduct an ablation study on our proposed blending Conditional Distillation on Geometry and Texture (CDGT). Our observations indicate that without Condition Updating Scheme in CDGT, the editing phase loses 3D-aware control, leading to unstable textures, and noticeable blurring. Without gradient combination, image synthesis in novel views  results in severe artifacts due to 3D degradation caused by mismatch between the underlying geometry and texture. The strategy proposed in TADA~\cite{tada} can partially alleviate this issue. However, the results still suffer from unstable texture and lead to hollow geometric structures. By incorporating gradient combination via random blend, we introduce randomness into the fusion process. This random blending boosts the resilience of edited results by preventing overfitting on specific patterns while enhancing its capability to handle challenging discrete attributes such as hats and glasses.

\begin{figure*}[t]
  \centering
  \includegraphics[width=0.97\linewidth]{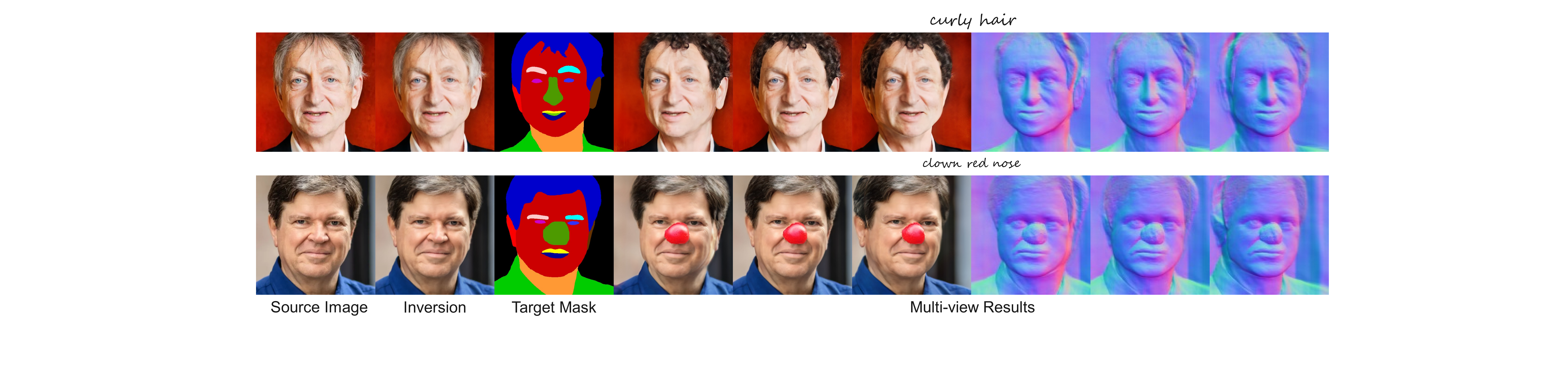}
  \caption{Real Portrait Editing. Our methods enable high-fidelity 3D inversion and make-guided text-based editing.}
  \label{fig:portrait}
\end{figure*}
\begin{figure*}[t]
  \centering
  \includegraphics[width=\linewidth]{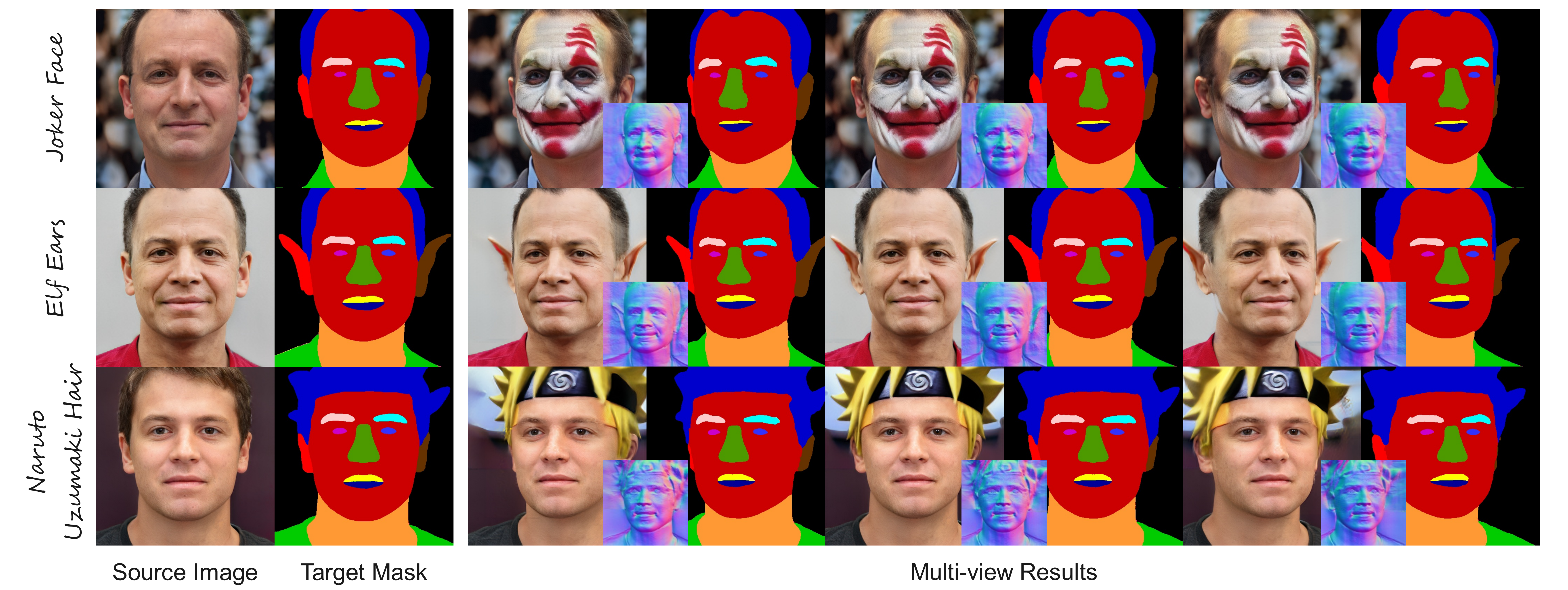}
  \caption{Out-of-Domain Editing. Our methods allow for precise modifications using out-of-domain textures.}
  \label{fig:outofdomain}
\end{figure*}

\begin{figure}[t]
  \centering
  \includegraphics[width=\linewidth]{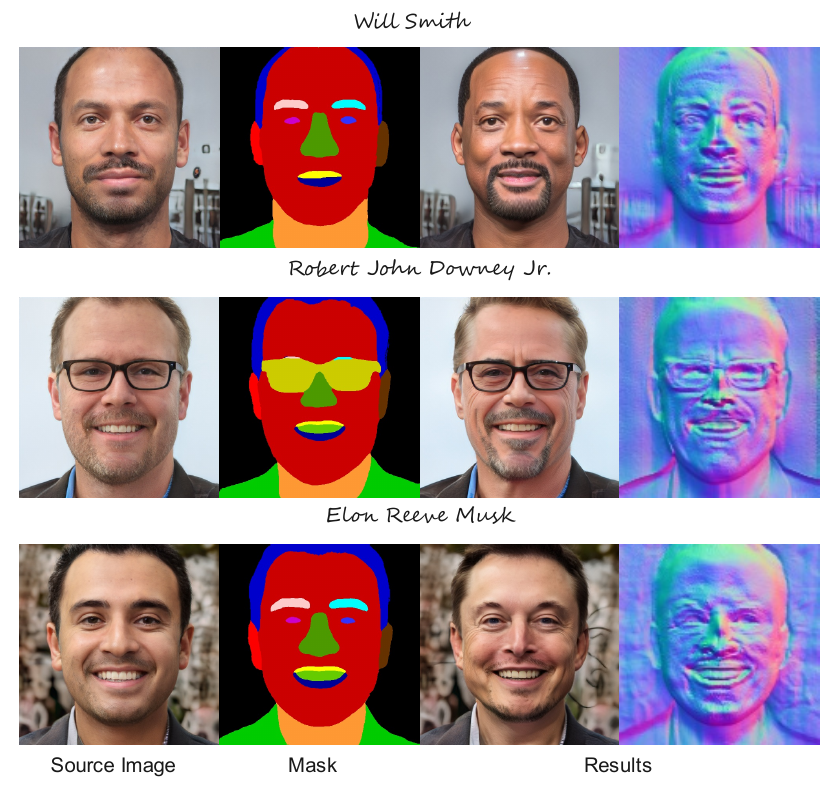}
  \caption{Face Swapping Editing. Our methods allow for face swapping with celebrities.}
  \label{fig:faceswap}
\end{figure}

\section{Applications}

\subsection{Real Portrait Editing}

Our MaTe3D allows for high-fidelity 3D inversion and mask-guided text-based editing on real portrait images. We start by projecting the input image into the latent space using PTI~\cite{pti}, and then use MaTe3D for mask-guided and text-based editing. The results are displayed in Fig.~\ref{fig:portrait}, demonstrating that our approach can achieve high-quality 3D portrait editing on real images.

\subsection{Out-of-Domain Editing}
Editing 3D faces with out-of-domain attributes is a challenging task due to the limitations of pre-trained generator training data. In our research, MaTe3D facilitates \textit{Out-of-Domain Editing} by creating a 3D face based on a target mask and text prompt. As shown in Fig.~\ref{fig:outofdomain}, our approach enables precise modifications with out-of-domain textures, while producing a convincing geometries for consistent visualization.

\subsection{Face Swapping}
Face swapping is a crucial research area in computer vision with extensive applications in entertainment industry~\cite{3dswap}. Our method can also perform face swap editing with celebrities. As shown in Fig.~\ref{fig:faceswap}, with a prompt of a celebrity, MaTe3D can transfer their face to the source image while keeping the remaining regions unchanged.

\section{Conclusion}
In this study, we propose MaTe3D, a novel editing pipeline to support mask-guided and text-based 3D-aware portrait editing 
 with high quality and stability. Extensive experiments demonstrated our significant performance over the-state-of-the-art methods. The proposed CatMask extends the mask-based editing domain from face to cat face, and provide an additional dataset to verify the feasibility, extensibility, and generalization of the model.

\minisection{Limitation} As previously mentioned, MaTe3D has a natural ability to enable mask-guided text-based editing and has surpassed its baselines. However, there are still some challenging cases that need to be addressed: 1) the image quality may slightly deteriorate when learning better geometry through SDFs, and 2) Our method is more time-consuming than baseline methods due to the optimized strategy. Much more efficient solutions need to be  explored in our future work.

\minisection{Ethical Concerns}
MaTe3D can reconstruct the representation of a given face and then launch mask-guided and text-based editing. A series of 3D-consistent images can thus be produced by taking advantage of MaTe3D, which means the great potential to empower areas like artistic creation and industrial design. However, the synthesized images can be wrongly identified by face recognition systems with high probability. Besides, downstream applications like single-view 3D inversion and mask-guided and text-based editing can be misused for generating edited imagery of real people. Such misusage put cyberspace in danger inevitably. Overall, the use of this technology needs to be more careful and better regulated.

\section*{Acknowledgments}
This work was supported by OpenBayes.

\bibliographystyle{IEEEtran}
\bibliography{final}


\end{document}